\documentclass[10pt,twocolumn,letterpaper]{article}

\usepackage{cvpr}
\usepackage{cvpr24}
\def\paperID{3}
\def\confName{CVPR}
\def\confYear{2024}

\newcommand{\bc}{\mathbf{c}}
\newcommand{\bC}{\mathbf{C}}
\newcommand{\bd}{\mathbf{d}}

\newcommand{\bx}{\mathbf{x}}
\newcommand{\bX}{\mathbf{X}}

\newcommand{\bI}{\mathbf{I}}

\newcommand{\bmu}{\boldsymbol\mu}
\newcommand{\bepsilon}{\boldsymbol\epsilon}
\newcommand{\bSigma}{\boldsymbol\Sigma}
\newcommand{\bbR}{\mathbb{R}}
\newcommand{\bbE}{\mathbb{E}}
\newcommand{\cC}{\mathcal{C}}
\newcommand{\cL}{\mathcal{L}}
\newcommand{\cN}{\mathcal{N}}
\newcommand{\cS}{\mathcal{S}}

\newcommand\blfootnote[1]{%
  \begingroup
  \renewcommand\thefootnote{}\footnote{#1}%
  \addtocounter{footnote}{-1}%
  \endgroup
}

\acrodef{cvae}[cVAE]{conditional Variational Autoencoder}
\acrodef{rl}[RL]{Reinforcement Learning}
\acrodef{mdp}[MDP]{Markov Decision Process}
\acrodef{bc}[BC]{Behavior Cloning}
\acrodef{hoi}[HOI]{Human-Object Interaction}
\acrodef{hsi}[HSI]{Human-Scene Interaction}
\acrodef{llm}[LLM]{Large Language Model}
\acrodef{coc}[CoC]{Chain of Contacts}
\acrodef{adm}[ADM]{Affordance Diffusion Model}
\acrodef{amdm}[AMDM]{Affordance-to-Motion Diffusion Model}
\acrodef{3dvl}[3D-VL]{3D Vision-Language}

\newcommand{\task}{\textit{language-guided human motion generation in 3D scenes}\xspace}

\title{Move as You Say, Interact as You Can:\\Language-guided Human Motion Generation with Scene Affordance\vspace{-9pt}}
\author{%
    Zan Wang$^{1,2}$, Yixin Chen$^{2}$, Baoxiong Jia$^{2}$, Puhao Li$^{2,3}$, Jinlu Zhang$^{2,4}$,\\
    Jingze Zhang$^{2,3}$, Tengyu Liu$^2$, Yixin Zhu$^{5\,\textrm{\Letter}}$, Wei Liang$^{1,6\,\textrm{\Letter}}$, Siyuan Huang$^{2\,\textrm{\Letter}}$
    \vspace{6pt}\\
    \small $^\textrm{\Letter}$ indicates corresponding authors\quad{}
    \small $^1$ School of Computer Science \& Technology, Beijing Institute of Technology\quad{}\\
    \small $^2$ National Key Laboratory of General Artificial Intelligence, BIGAI\quad{}
    \small $^3$ Dept. of Automation, Tsinghua University\\
    \small $^4$ CFCS, School of Computer Science, Peking University\quad{}
    \small $^5$ Institute for AI, Peking University\\
    \small $^6$ Yangtze Delta Region Academy of Beijing Institute of Technology, Jiaxing
    \vspace{6pt}\\
    \href{https://afford-motion.github.io}{https://afford-motion.github.io}
    \vspace{-15pt}
}

\begin{document}

\twocolumn[{%
\renewcommand\twocolumn[1][]{#1}%
\maketitle
\vspace{-15pt}
\begin{center}
    \centering
    \captionsetup{type=figure}
    \includegraphics[width=\linewidth]{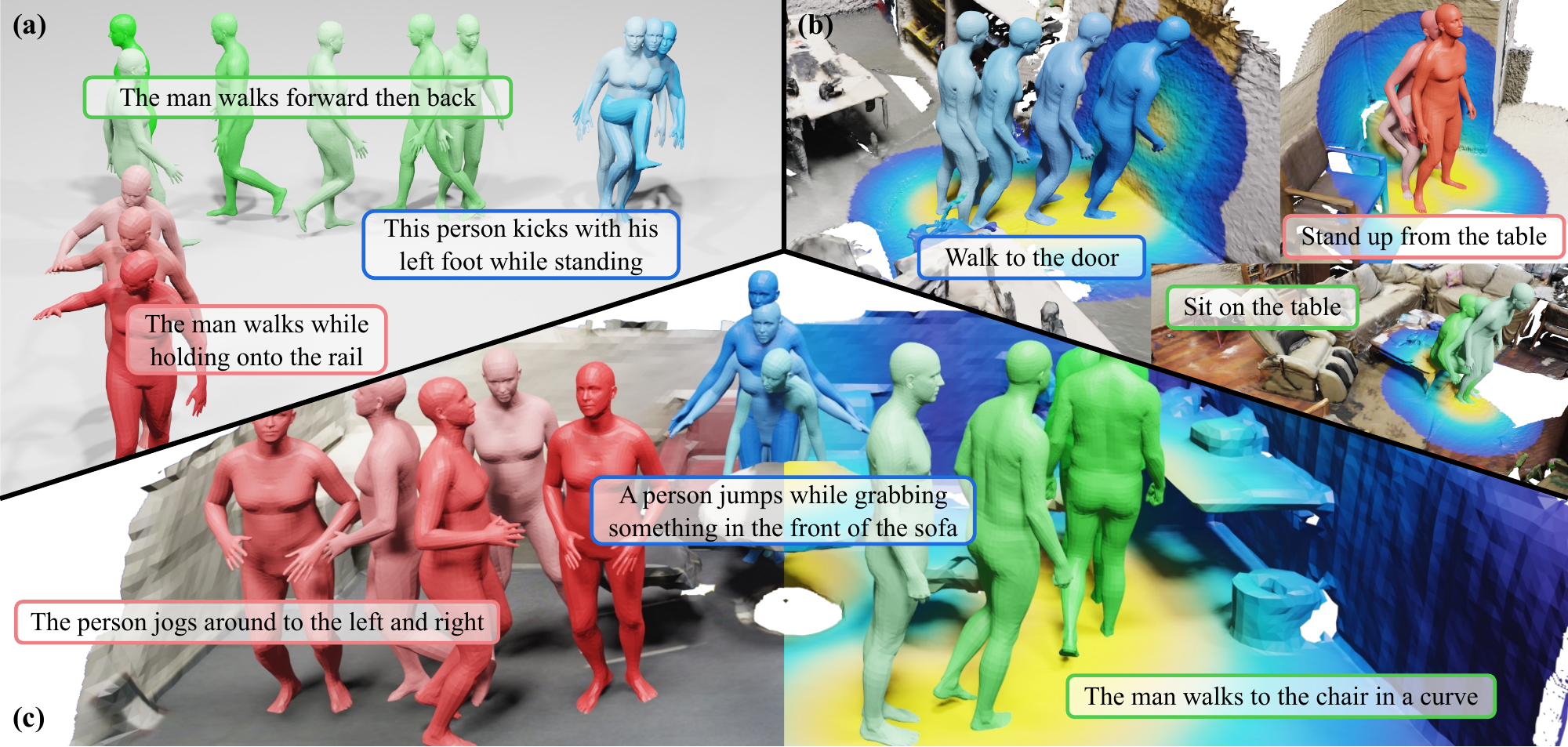}
    \captionof{figure}{\textbf{Language-guided human motion generation in 3D scenes via scene affordance.} Employing scene affordance as an intermediate representation enhances motion generation capabilities on benchmarks (a) HumanML3D and (b) HUMANISE, and significantly boosts the model's ability to generalize to (c) unseen scenarios.}
    \label{fig:teaser}
\end{center}
}]

\begin{abstract}\vspace{-12pt}
Despite significant advancements in text-to-motion synthesis, generating language-guided human motion within 3D environments poses substantial challenges. These challenges stem primarily from (i) the absence of powerful generative models capable of jointly modeling natural language, 3D scenes, and human motion, and (ii) the generative models' intensive data requirements contrasted with the scarcity of comprehensive, high-quality, language-scene-motion datasets. To tackle these issues, we introduce a \textbf{novel two-stage framework that employs scene affordance as an intermediate representation}, effectively linking 3D scene grounding and conditional motion generation. Our framework comprises an \ac{adm} for predicting explicit affordance map and an \ac{amdm} for generating plausible human motions. By leveraging scene affordance maps, our method overcomes the difficulty in generating human motion under multimodal condition signals, especially when training with limited data lacking extensive language-scene-motion pairs. Our extensive experiments demonstrate that our approach consistently outperforms all baselines on established benchmarks, including HumanML3D and HUMANISE. Additionally, we validate our model's exceptional generalization capabilities on a specially curated evaluation set featuring previously unseen descriptions and scenes.
\end{abstract}

\section{Introduction}

Prior efforts in the field have investigated the integration of diverse modalities, such as textual descriptions~\citep{ahuja2019language2pose,ghosh2021synthesis,guo2022generating,petrovich2022temos,tevet2022motionclip,athanasiou2022teach,zhang2022motiondiffuse,tevet2023human,zhang2023generating,chen2023executing,athanasiou2023sinc}, audio signals~\citep{lee2019dancing,li2021audio2gestures,yi2023generating}, and 3D scenes~\citep{wang2021synthesizing,wang2021scene,hassan2021stochastic,wang2022towards,araujo2023circle,huang2023diffusion} for guiding human motion generation. The significant strides in single-modality conditioned motion generation have been complemented by the introduction of \ac{hsi} through language descriptions by \citet{wang2022humanise}, highlighting the demand for controllable motion generation in diverse applications such as animation synthesis~\citep{holden2016deep}, film production~\citep{thorne2004motion}, and synthetic data generation~\citep{zhang2020generating,zhao2022compositional}. However, the task of effectively generating semantically driven and scene-aware motions remains daunting due to two principal challenges.

The first challenge entails ensuring that generated motions are descriptive-faithful, physically plausible within the scene, and accurately grounded in specific locations. Though direct application of conditional generative models like \ac{cvae}~\citep{petrovich2021action,wang2021synthesizing,wang2022humanise} and conditional diffusion models~\citep{zhang2022motiondiffuse,tevet2023human,chen2023executing,huang2023diffusion} has been attempted, the inherent complexity of marrying 3D scene grounding with conditional motion generation presents a significant obstacle. This complexity impedes the model's ability to generalize across various scenes and descriptions, making it challenging to adapt specific motions (\eg, \textit{``lie down on the bed''}) to analogous actions in new contexts (\eg, \textit{``lie down on the floor''}) within unfamiliar 3D environments.

The second challenge arises from the generative models' dependency on large volumes of high-quality paired data. Existing \ac{hsi} datasets~\citep{hassan2019resolving,cao2020long,araujo2023circle} lack in both motion quality and diversity, featuring a limited number of scene layouts and, most critically, devoid of \ac{hsi} descriptions. Although the HUMANISE dataset~\citep{wang2022humanise} attempts to address this gap, it is constrained by a narrow scope of action types and the use of fixed-form utterances, limiting the generation of diverse \acp{hsi} from varied and free-form language descriptions.

In response to these challenges, we propose to utilize the \textbf{scene affordance maps as an intermediate representation}, as depicted in \cref{fig:teaser}. This representation is calculated from the distance field between human skeleton joints and the scene's surface points. The use of the affordance map presents two primary benefits for the generation of language-guided motion in 3D environments. First, it precisely delineates the region grounded in the language description, thereby significantly enhancing the 3D scene grounding essential for motion generation, even in scenarios characterized by limited training data availability. Second, the affordance map, rooted in distance measurements, provides a sophisticated understanding of the geometric interplay between scenes and human motions. This understanding aids in the generation of \ac{hsi} and facilitates the model's ability to generalize across unique scene geometries.

Expanding upon this intermediate representation, we propose a novel two-stage model aimed at seamlessly integrating the 3D scene grounding with the language-guided motion generation. The first stage involves the development of an \textbf{\acf{adm}}, which employs the Perceiver architecture~\citep{jaegle2021perceiver,jaegle2021perceiverio} to predict an affordance map given a specific 3D scene and description. The second stage introduces an \textbf{\acf{amdm}}, comprising an affordance encoder and a Transformer backbone, to synthesize human motions by considering both the language descriptions and the affordance maps derived in the first stage.

We conduct extensive evaluations on established benchmarks, including HumanML3D~\citep{guo2022generating} and HUMANISE~\citep{wang2022humanise}, demonstrating superior performance in text-to-motion generation tasks and highlighting our model's advanced generalization capabilities on a specially curated evaluation set featuring unseen language descriptions and 3D scenes. These results underscore the utility of our approach in harnessing scene affordances for enriched 3D scene grounding and enhanced conditional motion generation.

Our contributions are summarized as follows:
\begin{itemize}[leftmargin=*,nolistsep,noitemsep]
    \item We introduce a novel two-stage model that incorporates scene affordance as an intermediate representation, bridging the gap between 3D scene grounding and conditional motion generation, and facilitating language-guided human motion synthesis in 3D environments.
    \item Through extensive quantitative and qualitative evaluations, we demonstrate our method's superiority over existing motion generation models across the HumanML3D and HUMANISE benchmarks.
    \item Our model showcases remarkable generalization capabilities, achieving impressive performance in generating human motions for novel language-scene pairs, despite the limited availability of language-scene-motion datasets.
\end{itemize}

\section{Related Work}

\subsection{Language, Human Motion, and 3D Scene}

We seek to bridge the modalities of language, human motion, and 3D scenes, an area where prior research has often focused on combining just two of these elements. In the realm of \ac{3dvl}, tasks such as 3D object grounding~\citep{achlioptas2020referit3d,chen2021yourefit,thomason2022language,zhang2023multi3drefer,zhu20233d,jia2024sceneverse}, reasoning~\citep{das2018embodied,ye20223d,azuma2022scanqa,ma2022sqa3d}, and captioning~\citep{chen2021scan2cap,yuan2022x,chen2021d3net,chen2023end,huang2023embodied} have intersected language with 3D scenes. Recent advancements in this area have focused on enhancing open-vocabulary scene understanding by integrating features from foundational models like CLIP~\citep{radford2021learning} into 3D scene analysis~\citep{lerf2023,jatavallabhula2023conceptfusion,peng2023openscene,takmaz2023openmask3d}. The interaction between language and human motion has been explored through efforts to guide motion generation with semantic cues, including text-to-motion~\citep{ahuja2019language2pose,ghosh2021synthesis,guo2022generating,petrovich2022temos,athanasiou2023sinc} and action-to-motion synthesis~\citep{guo2020action2motion,petrovich2021action}.

Existing \ac{hsi} works focus on populating static human figures into 3D scenes~\citep{chen2019holistic,zhang2020generating,hassan2021populating,chen2023detecting} and generating temporal human motions within these contextual environments~\citep{wang2021synthesizing,wang2021scene,wang2022towards,huang2023diffusion,jiang2023full}. A growing body of research~\citep{ling2020character,peng2021amp,peng2022ase,zhang2022wanderings,zhao2023synthesizing,cui2024anyskill} has aimed at creating policies for continuous motion synthesis in virtual spaces, treating the challenge as a \ac{rl} task. The pioneering works of \citet{zhao2022compositional} and \citet{wang2022humanise} ventured into the simultaneous modeling of language, 3D scenes, and human motion, integrating semantics (\eg, action labels and descriptive language) into the generation of \ac{hsi}, requiring interactions to be both physically plausible and semantically consistent. Following this, \citet{xiao2023unified} leveraged a \ac{llm} to convert language prompts into sub-task plans, represented as \ac{coc}, to facilitate motion planning within 3D scenes.

In our contribution, we present a novel two-stage framework that employs scene affordance~\citep{gibson1977theory} as an intermediary to effectively bridge 3D scene grounding with conditioned motion generation. This approach not only enhances multimodal alignment but also improves the generative model's ability to generalize across scenarios, even when trained on the limited paired data available in current datasets~\citep{guo2022generating,hassan2019resolving,wang2022humanise}.

\subsection{Conditional Human Motion Generation}

The past few years have marked significant advancements in the domain of human motion modeling conditioned on diverse signals~\citep{zhu2023human}, including past motion~\citep{li2018convolutional,barsoum2018hp,yuan2020dlow,cao2020long,xie2021physics,mao2022contact}, audio~\citep{lee2019dancing,li2021audio2gestures,yi2023generating}, action labels~\citep{guo2020action2motion,petrovich2021action}, natural language descriptions~\citep{ahuja2019language2pose,ghosh2021synthesis,guo2022generating,petrovich2022temos,tevet2022motionclip,athanasiou2022teach,zhang2022motiondiffuse,tevet2023human,zhang2023generating,chen2023executing,athanasiou2023sinc}, objects~\citep{chao2021learning,taheri2022goal,zhang2022couch,ghosh2023imos,li2023object}, and 3D scenes~\citep{wang2021synthesizing,wang2021scene,hassan2021stochastic,wang2022towards,araujo2023circle,huang2023diffusion,jiang2024scaling}. These approaches, predominantly designed for single-modal conditioning, encounter difficulties in scenarios necessitating the simultaneous consideration of both scene and language cues. For example, methods that seek to align the conditional signal's latent space with that of human motions~\citep{ahn2018text2action,tevet2022motionclip,petrovich2022temos,athanasiou2023sinc} struggle in this intricate context due to the distinct and complementary nature of 3D scenes and language descriptions in motion generation. The former provides spatial boundaries while the latter offers semantic direction, rendering direct alignment approaches less effective.

Moreover, attempts at directly learning the conditional distribution with models such as \ac{cvae}~\citep{petrovich2021action,wang2021synthesizing,wang2022humanise} and diffusion models~\citep{zhang2022motiondiffuse,tevet2023human,chen2023executing} often lead to suboptimal outcomes. This is attributed to the complex entanglement of the joint distribution across the three modalities, which complicates the development of an efficient multimodal embedding space, particularly when data is scarce. In response, our research proposes the utilization of scene affordance as an intermediate representation. This strategy aims to simplify the process of generating motion under multiple conditions, thereby enhancing the model's capacity to interpret and generate multimodal \ac{hsi} more effectively.

\subsection{Scene Affordance}

The concept of ``affordance,'' initially introduced by \citet{gibson1977theory}, describes the potential actions that the environment offers for interaction. Early investigations into affordances primarily focused on understanding scenes and object affordances through 2D observations~\citep{grabner2011makes,gupta2011from,zhu2014reasoning,koppula2014physically,zhu2016inferring,wang2017binge,fang2018demo2vec,li2019putting,nagarajan2019grounded,kulal2023putting}. Transitioning to 3D, initial \ac{hsi} research implicitly incorporated affordances in scene understanding~\citep{wang2021synthesizing,wang2021scene,wang2022towards,wang2022humanise}, with more recent work exploring explicit 3D visual affordances~\citep{zhu2016inferring,nagarajan2020learning,deng20213d}. These advancements often represent affordances as contact maps for grasping~\citep{kokic2020learning,jiang2021hand,li2023gendexgrasp,wu2023learning,li2024grasp} and scene-conditioned motion synthesis~\citep{zhang2020generating,zhang2020place,hassan2021populating,wang2021synthesizing,wang2021scene}. In our approach, we redefine the affordance map as a generalized distance field between human skeleton joints and surface points of 3D scenes. This model first refines the affordance map using the provided 3D scene and language description. It then utilizes the refined affordance map's grounding and geometric information to improve the subsequent conditional motion generation.

\begin{figure*}[t!]
    \centering
    \includegraphics[width=\linewidth]{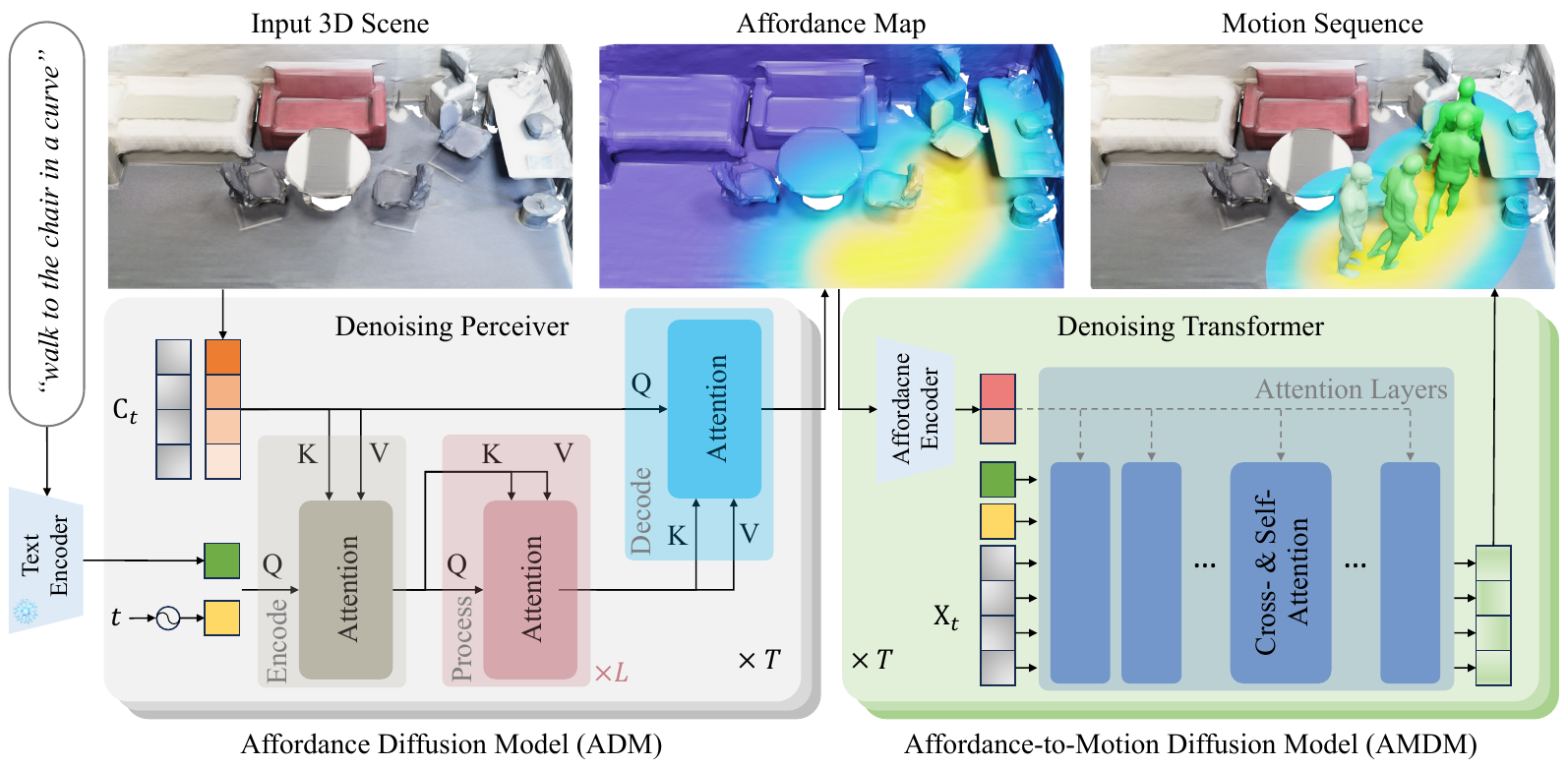}
    \caption{\textbf{Overview of our method.} To generate language-guided human motions in 3D scenes, our framework first predicts the scene affordance map in accordance with the language description using \acf{adm}. Next, it generates interactive human motions with \acf{amdm} conditioned on the predicted affordance map.}
    \label{fig:pipeline}
\end{figure*}

\section{Preliminaries}

\paragraph{Diffusion Model}

Diffusion models~\citep{sohl2015deep,ho2020denoising,song2019generative} are a class of generative models that operate through an iterative denoising process to learn and sample data distributions. They include a forward process and a reverse process.

The forward process starts with the real data $\bX_0$ at step 0, iteratively adds Gaussian noise $\bepsilon_t$, and converts $\bX_0$ to $\bX_t$ over $t$ steps in a Markovian manner. The one-step forward process can be described as
$
    q(\bX_t \mid \bX_{t-1}) = \cN(\bX_t; \sqrt{1 - \beta_t}\bX_{t-1}, \beta_t\bI),
$
where $\left\{\beta_t \in (0, 1)\right\}_{t=1}^T$ is the pre-defined variance schedule. When $t \to \infty$, $\bX_t$ is equivalent to an isotropic Gaussian distribution. The entire forward process is given by
$
    q(\bX_{1:T} \mid \bX_0) = \prod_{t=1}^Tq(\bX_t | \bX_{t-1}),
$
where $T$ is the total number of diffusion steps.

In the reverse process, the diffusion model learns to gradually remove noise for sampling from the Gaussian distribution $\bX_T$:
$
    p_\theta(\bX_{0:T}) = p(\bX_T)\prod_{t=1}^T p_\theta (\bX_{t-1} \mid \bX_t),
    p_\theta(\bX_{t-1} \mid \bX_t) = \cN(\bX_{t-1}; \bmu_\theta(\bX_t, t), \bSigma_\theta(\bX_t, t)),
$
where $\bX_T \sim \cN(\textbf{0}, \bI)$, $\bmu_\theta$ and $\bSigma_\theta$ are estimated by the models with learnable parameters $\theta$.

For learning a conditional distribution $p_\theta(\bX_0 \mid \cC)$, the diffusion model can be adapted to include condition $\cC$ in the reverse process:
$
    p_\theta(\bX_{t-1} \mid \bX_t, \cC) = \cN(\bX_{t-1}; \bmu_\theta(\bX_t, t, \cC), \bSigma_\theta(\bX_t, t, \cC)).
$

\paragraph{Problem Definition}

We tackle the task of \task. The 3D scene is represented as an RGB point cloud $\cS \in \bbR^{N \times 6}$, while the language description is denoted as $\cL = \left[ w_1, w_2, \cdots, w_M\right]$, comprising $M$ tokenized words. Our objective is to generate motion sequences $\bX = \left\{\bx_i\right\}_{i=1}^F$ that are both physically plausible and semantically consistent with the given descriptions, where each sequence consists of $F$ frames. Diverging from the redundant motion representation used by \citet{guo2022generating}, we parameterize the per-frame human pose using the body joint positions of the SMPL-X body model~\citep{pavlakos2019expressive}, specifically $\bx_i \in \bbR^{J \times 3}$, with $J$ representing the total number of joints utilized. For visualization purposes, these poses are converted into body meshes by optimizing the SMPL-X parameters based on joint positions. Please refer to \cref{app:sec:joints2mesh} for additional details on the optimization process.

\section{Method}

We propose a novel two-stage model for generating plausible human motions conditioned on the 3D scene and language descriptions. \cref{fig:pipeline} illustrates the model's framework. The first stage introduces an \acf{adm} to generate language-grounded affordance maps. The second stage takes input as the generated affordance map and the language description to synthesize plausible human motions from Gaussian noise via the proposed \acf{amdm}.

\subsection{Affordance Map}

The affordance map serves as an intermediate representation that abstracts essential details of a 3D indoor scene to support generalization, accurately ground interaction regions, and preserve vital geometric information. In this work, we derive such an affordance map from the distance field between the points in a 3D scene $\cS$ and the human skeleton joints across a motion sequence $\bX = \left\{\bx_i\right\}_{i=1}^F$. We calculate the $\ell_2$ distance between each scene point and the skeleton joints per frame, resulting in a per-frame distance field $\bd \in \bbR^{N \times J}$; $\bd(n, j)$ measures the distance between the $n$-th scene point and the $j$-th skeleton joint. Following \citet{mao2022contact}, we transform this distance field into a normalized distance map $\bc \in \bbR ^{N \times J}$:
\begin{equation}
    \small
    \bc(n, j)=\exp\left(-\frac{1}{2}\frac{\bd(n, j)}{\sigma^2}\right),
    \label{eq:dist_to_affor}
\end{equation}
where $\sigma$ is a constant normalizing factor. This operation assigns higher weights to points closer to the joints, thereby aiding in stabilizing the training procedure.

To compute the affordance map $\bC$, we employ a max-pooling operation over the temporal dimension of the per-frame distance fields:
\begin{equation}
    \small
    \bC = \texttt{max-pool}(\bc_{1}, \bc_{2}, \ldots, \bc_{F}).
\end{equation}
The resulting paired data is denoted as $(\bC, \bX, \cS, \cL)$, with $\cS$ and $\cL$ representing the scene's point cloud and the associated language description, respectively.

\subsection{Affordance Diffusion Model}

To learn the distribution of language-grounded affordance maps, we introduce an \acf{adm} designed to process the 3D scene point cloud $\cS$ and the corresponding language description $\cL$, generating an affordance map $\bC$. This process is formalized as follows:
\begin{equation}
    \small
    p_\theta(\bC_{0:T} \mid \cS, \cL) = p(\bC_T) \prod_{t=1}^T p_\theta(\bC_{t-1} \mid \bC_{t}, \cS, \cL).
\end{equation}
As depicted in \cref{fig:pipeline}, \ac{adm}'s architecture is based on the Perceiver~\citep{jaegle2021perceiver,jaegle2021perceiverio}, leveraging an attention mechanism to efficiently extract point-wise features.

The Perceiver backbone within \ac{adm} consists of three primary components: an \textit{Encode} block, a \textit{Process} block, and a \textit{Decode} block. Initially, the \textit{Encode} utilizes an attention module to encode the extracted point features along with the noisy affordance map, termed as input features. We denote the concatenation of the language feature and diffusion step embeddings as the latent features. In this configuration, the input features act as the attention module's key and value, with the latent features acting as the query. Next, the \textit{Process} block refines the latent features through multiple self-attention layers. Finally, the \textit{Decode} block employs another attention module, allowing the input features to attend to the updated latent features, thereby achieving refined per-point feature refinement. We forward these per-point feature vectors into a linear layer for further processing. Contrary to approaches that predict the added noise $\bepsilon_t$, our model directly estimate the input signal \citep{ramesh2022hierarchical,tevet2023human}, allowing the end-to-end training of \ac{adm}, denoted as $G_\theta$, with a simple objective:
\begin{equation}
    \small
    L_{\text{MSE}} = \bbE_{\bC_0, t}\left[\norm{\bC_0 - G_\theta(\bC_t, t, \cS, \cL)}_2^2\right].
\end{equation}

\subsection{Affordance-to-Motion Diffusion Model}

In the subsequent stage, our framework employs an \acf{amdm} to generate plausible human motions, leveraging both the language descriptions and the previously generated affordance maps:
\begin{equation}
    \small
    p_\phi(\bX_{0:T} \mid \bC, \cS, \cL) = p(\bX_T)\prod_{t=1}^{T} p_{\phi}(\bX_{t-1} \mid \bX_{t}, \bC, \cS, \cL).
\end{equation}
The architecture of the model is illustrated in \cref{fig:pipeline}. The \ac{amdm} comprises an encoder specifically for the affordance map and a Transformer backbone that integrates multimodal features to facilitate motion generation. Utilizing a Point Transformer architecture~\citep{zhao2021point}, the affordance map encoder extracts feature maps of varying cardinalities, which are further processed by U-net decoder layers; the Transformer backbone stacks self-attention and cross-attention layers. We concatenate the noisy motion sequence with language features and diffusion timestep embeddings and forward this concatenation to the Transformer backbone. In each cross-attention layer, the concatenation attends to the affordance features to fuse multimodal information. A linear layer finally maps the fused features into the motion space.

Similar to \ac{adm}, we train the \ac{amdm}, denoted as $G_\phi$, by optimizing a mean squared error objective:
\begin{equation}
    \small
    L_{\text{MSE}} = \bbE_{\bX_0, t}\left[\norm{\bX_0 - G_\phi(\bX_t, t, \bC, \cS, \cL)}_2^2\right].
\end{equation}

\subsection{Implementation Details}

In our implementation, we use a frozen \textit{CLIP-ViT-B/32} to extract text features in both stages. The normalization factor $\sigma$ is set to 0.8. The Transformer models are constructed using the native PyTorch implementation. Both \ac{adm} and \ac{amdm} undergo training to convergence using the AdamW optimizer with a fixed learning rate of $10^{-4}$. For the training of \ac{adm}, we leverage 2 NVIDIA A100 GPUs, assigning a batch size of 64 per GPU. The training of \ac{amdm} is conducted on 4 NVIDIA A100 GPUs, with a batch size of 32 per GPU. Refer to \cref{app:sec:implementation} for further implementation details.

\begin{table*}[ht!]
    \centering
    \small
    \setlength{\tabcolsep}{3pt}
    \caption{\textbf{Quantitative results of generation on HumanML3D.} ``Real'' denotes the results computed with GT motions. ``$\rightarrow$'' indicates metrics that are better when closer to ``Real'' distribution. Our model uses Perceiver in \ac{adm} and encoder-based architecture in \ac{amdm}.}
    \label{tab:humanml_quan}
        \begin{tabular}{cccccccc}%
            \toprule
            \multirow{2}{*}{Model} & \multicolumn{3}{c}{R-Precision $\uparrow$} & \multirow{2}{*}{FID $\downarrow$} & \multirow{2}{*}{MultiModal Dist. $\downarrow$} & \multirow{2}{*}{Diversity $\rightarrow$} & \multirow{2}{*}{MultiModality $\uparrow$}\\
            \cline{2-4} & Top 1 &  Top 2 & Top 3 & \\
            \midrule
            Real                                            & $0.511^{\pm.003}$          & $0.703^{\pm.003}$          & $0.797^{\pm.002}$          & $0.002^{\pm.000}$          & $2.974^{\pm.008}$          & $9.503^{\pm.065}$          & -                           \\
            \midrule
            Language2Pose \citep{ahuja2019language2pose}    & $0.246^{\pm.002}$          & $0.387^{\pm.002}$          & $0.486^{\pm.002}$          & $11.02^{\pm.046}$          & $5.296^{\pm.008}$          & $7.676^{\pm.058}$          & -                           \\
            T2M \citep{guo2022generating}                   & $\mathbf{0.457^{\pm.002}}$ & $\mathbf{0.639^{\pm.003}}$ & $\mathbf{0.740^{\pm.003}}$ & $1.067^{\pm.002}$          & $\mathbf{3.340^{\pm.008}}$ & $9.188^{\pm.002}$          & $2.090^{\pm.083}$           \\
            MDM \citep{tevet2023human}                      & $0.319^{\pm.005}$          & $0.498^{\pm.004}$          & $0.611^{\pm.007}$          & $0.544^{\pm.044}$          & $5.566^{\pm.027}$          & $\mathbf{9.559^{\pm.086}}$ & $2.799^{\pm.072}$           \\
            Ours                                            & $0.341^{\pm.010}$          & $0.514^{\pm.016}$          & $0.625^{\pm.011}$          & $\mathbf{0.352^{\pm.109}}$ & $5.455^{\pm.073}$          & $9.772^{\pm.117}$          & $\mathbf{2.835^{\pm.075}}$  \\
            \midrule
            $\text{MDM}^\dag$ \citep{tevet2023human}        & $0.418^{\pm.005}$          & $0.604^{\pm.005}$          & $0.707^{\pm.004}$          & $0.489^{\pm.025}$          & $3.631^{\pm.023}$          & $\mathbf{9.449^{\pm.066}}$ & $\mathbf{2.873^{\pm.111}}$  \\
            $\text{Ours}^\dag$                              & $\mathbf{0.432^{\pm.007}}$ & $\mathbf{0.629^{\pm.007}}$ & $\mathbf{0.733^{\pm.006}}$ & $\mathbf{0.352^{\pm.109}}$ & $\mathbf{3.430^{\pm.061}}$ & $9.825^{\pm.159}$          & $2.835^{\pm.075}$           \\
            \bottomrule
        \end{tabular}%
\end{table*}

\begin{table*}[t!]
    \centering
    \small
    \setlength{\tabcolsep}{3pt}
    \caption{\textbf{Quantitative results of human motion generation on HUMANISE dataset.} \textbf{Bold} indicates the best result.}
    \label{tab:humanise_quan}
        \begin{tabular}{ccccccc}%
            \toprule
            Model &  goal dist.$\downarrow$ & APD$\uparrow$  & contact$\uparrow$ & non-collision$\uparrow$ & quality score$\uparrow$ & action score$\uparrow$ \\
            \midrule
            \ac{cvae} \citep{wang2022humanise} & $0.422^{\pm.011}$          & $4.094^{\pm.013}$          & $84.06^{\pm.716}$          & $\mathbf{99.77^{\pm.004}}$ & $2.25\pm1.26$ & $3.66\pm1.38$ \\
            one-stage @ Enc                    & $0.326^{\pm.013}$          & $\mathbf{5.510^{\pm.019}}$ & $76.11^{\pm.684}$          & $99.71^{\pm.014}$          & $2.60\pm1.24$ & $3.88\pm1.32$ \\
            one-stage @ Dec                    & $0.185^{\pm.014}$          & $4.063^{\pm.020}$          & $86.43^{\pm.845}$          & $99.76^{\pm.006}$          & $3.09\pm1.34$ & $4.18\pm1.16$ \\
            \midrule
            Ours @ Enc                         & $\mathbf{0.156^{\pm.006}}$ & $2.597^{\pm.008}$          & $95.86^{\pm.323}$          & $99.69^{\pm.007}$          & $3.46\pm1.15$ & $\mathbf{4.47\pm0.84}$ \\
            Ours @ Dec                         & $\mathbf{0.156^{\pm.006}}$ & $2.459^{\pm.009}$          & $\mathbf{96.04^{\pm.298}}$ & $99.70^{\pm.005}$          & $\mathbf{3.55\pm1.19}$ & $4.44\pm0.85$ \\
            \bottomrule
        \end{tabular}%
\end{table*}

\section{Experiments}

To demonstrate the efficacy of our methods, we conducted evaluations using the HumanML3D~\citep{guo2022generating}, HUMANISE~\citep{wang2022humanise}, and a uniquely compiled evaluation set specifically curated for examining the generalization capability.

\subsection{Datasets}

We evaluate our model on HumanML3D~\citep{guo2022generating}, a modern text-to-motion dataset derived from annotating AMASS~\citep{mahmood2019amass} motion sequences with sequential-level descriptions. As HumanML3D lacks 3D scenes, we augment it by adding a floor to support the training and evaluation of our two-stage model. We use the original motion representation and train-test splits in the task setting.

We also evaluate our model on HUMANISE~\citep{wang2022humanise}, distinguished as the first extensive and semantic-rich \ac{hsi} dataset that aligns motion sequences from AMASS with the 3D scene from ScanNet~\citep{dai2017scannet}. The synthesized results are automatically annotated with descriptions from Sr3D~\citep{achlioptas2020referit3d}. We exclude spatially referring descriptions and segment scenes into chunks while retaining the original motions and splits.

To probe the model's generalization prowess, we curate a novel evaluation set that comprises 16 scenes from diverse sources, including ScanNet~\citep{dai2017scannet}, PROX~\citep{hassan2019resolving}, Replica~\citep{straub2019replica}, and Matterport3D~\citep{chang2017matterport3d}, along with 80 \ac{hsi} descriptions crafted by Turkers. Furthermore, we construct a training set that connects language, 3D scene, and motion by incorporating data from HumanML3D, HUMANISE, and PROX. We leverage the annotations to unify the representation as joint positions across different datasets; we augment the HumanML3D by randomly positioning furniture~\citep{uy2019revisiting} around the motion to boost 3D scene awareness. This consolidated dataset comprises 63,770 \acp{hsi}, with 48,470 featuring language annotations. Refer to \cref{app:sec:novel_eval} for more details.

\subsection{Metrics and Baselines}

\paragraph{Metrics}

For the evaluation on \textbf{HumanML3D}, we adopt the metrics proposed by \citet{guo2022generating}, including \textit{Diversity}, measuring the variation within generated motions; \textit{MultiModality}, quantifying the average variation relative to text descriptions; \textit{R-Precision} and \textit{Multimodal-Dist}, assessing the relevance between generated motions and language descriptions; and \textit{FID}, evaluating the discrepancy between the distributions of generated results and the original dataset.
On \textbf{HUMANISE}, we follow the evaluation protocol of \citet{wang2022humanise} and \citet{zhang2020generating}, utilizing metrics of \textit{goal dist.} to determine grounding accuracy, \textit{Average Pairwise Distance (APD)} for the diversity of motions, and physics-based metrics like \textit{contact} and \textit{non-collision} scores. Human perceptual studies further evaluate the \textit{quality} and \textit{action} score of the generated motions.
\blfootnote{``$\dag$'' indicates adjustments following bug fixes in the evaluation code, detailed at \href{https://github.com/GuyTevet/motion-diffusion-model/issues/182}{https://github.com/GuyTevet/motion-diffusion-model/issues/182}.}
To evaluate \textbf{\ac{adm}}, we further introduce three grounding metrics, \ie, \textit{min dist.}, \textit{pelvis dist.}, and \textit{all dist.}, to quantify the accuracy of the affordance map in guiding interactions, based on distances from the joints to target objects within the scene.
We also employ these metrics on the \textbf{novel evaluation set}. Due to unique motion representations, we retrain the motion and text feature extractors as follows \citet{guo2022generating} for consistent metric calculations.
All evaluations are conducted five times to ensure robustness, with a $95\%$ confidence interval indicated by $\pm$. For \textit{quality} and \textit{action} scores, mean and standard deviation are reported.

\paragraph{Baselines}

For evaluations on \textbf{HumanML3D}, we include the following baselines: Language2Pose~\citep{ahuja2019language2pose}, T2M~\citep{guo2022generating}, and MDM~\citep{tevet2023human}. For \textbf{HUMANISE}, we utilize the \ac{cvae}-based approach by \citet{wang2022humanise}, hereafter referred to as \textit{cVAE}.
To evaluate \textbf{scene affordances} and our \textbf{two-stage model} architecture, we implement an \textit{one-stage} diffusion model variant that directly processes the scene point cloud, bypassing the affordance map generation; this variant replicates the \ac{amdm}'s architecture. Moreover, we examine an encoder adaptation of \ac{amdm} that integrates the concatenated features from the three modalities (human motion, affordance map, and language description) directly into self-attention layers, serving additional baselines. The designations \textit{@ Enc} and \textit{@ Dec} refer to encoder and decoder variants, respectively.
For \textbf{affordance map generation} in the first stage, we explore two additional architectural variations of \ac{adm}, MLP and Point Transformer, to further understand their impact on performance. Further details of baseline models' architecture are available in \cref{app:sec:architecture}.

\begin{figure*}[t!]
    \centering
    \includegraphics[width=\linewidth]{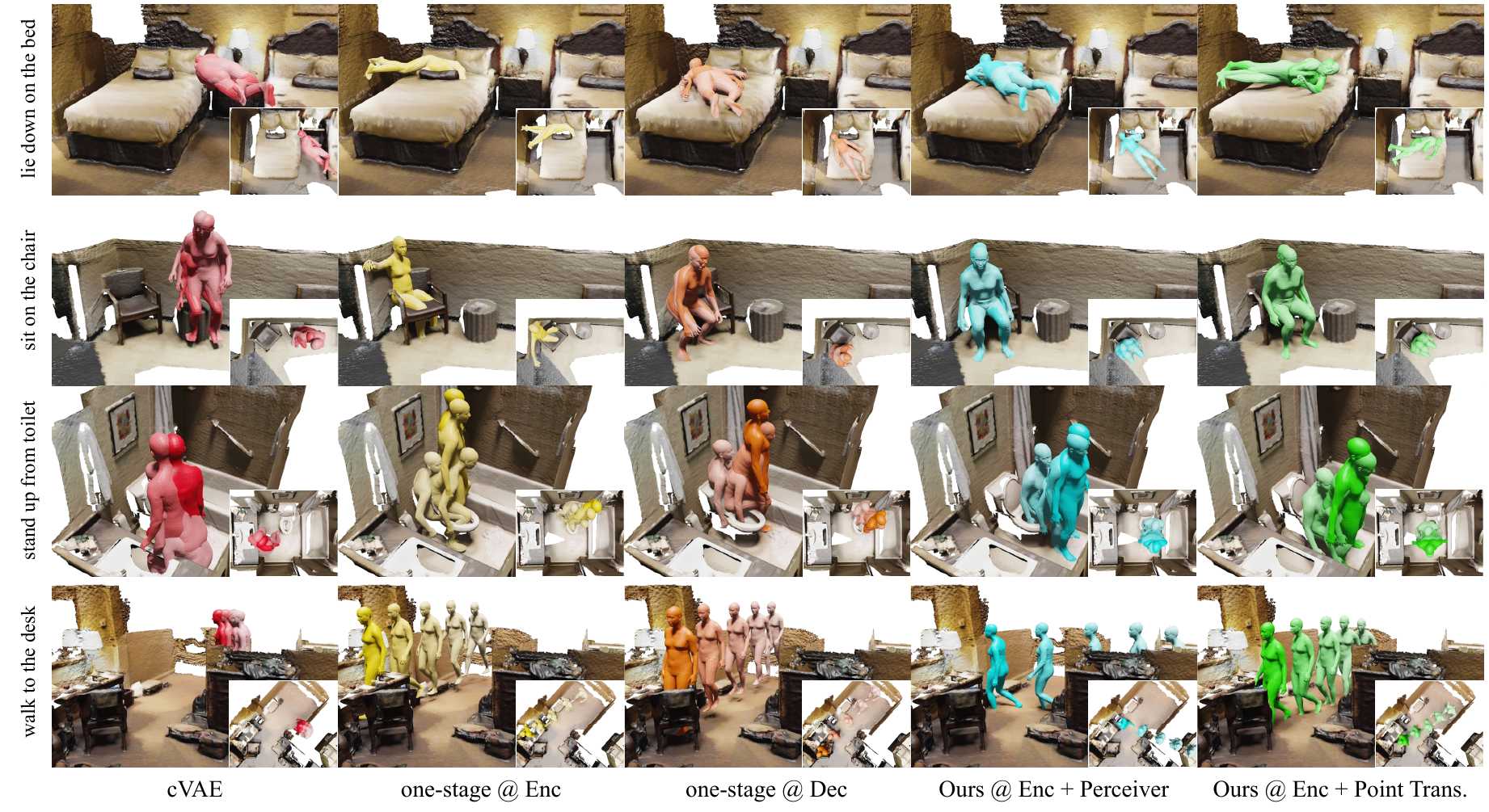}
    \caption{\textbf{Qualitative results on HUMANISE dataset.} The bottom-right figure provides a top-down view. Zoom in for better visualization.}
    \label{fig:humanise_qual}
\end{figure*}

\subsection{Results on HumanML3D}

\cref{tab:humanml_quan} showcases the quantitative results on HumanML3D, where our method notably excels in the \textit{FID} metric, outperforming all baselines. Specifically, against MDM~\citep{tevet2023human}, our method demonstrates enhanced performance in \textit{R-Precision}, \textit{FID}, and \textit{MultiModal Dist.}, while preserving a comparable level of diversity, even in the absence of auxiliary geometric losses. Given that MDM stands as a leading diffusion model in motion generation with a Transformer backbone similar to ours, these findings underscore the benefits of integrating scene affordance into text-to-motion synthesis by enriching movement details such as joint trajectories, evidenced even with a simple floor augmentation to the language-motion dataset. \cref{app:sec:more_qual_results} provides qualitative results of the predicted affordance maps and generated motions.

\subsection{Results on HUMANISE}

\paragraph{Quantitative Results}

Quantitative evaluations presented in \cref{tab:humanise_quan} affirm our method's capability in producing high-fidelity human motion sequences that are well-grounded conditioned on scenes and language instructions, outperforming both the \textit{cVAE} and one-stage diffusion model baselines. Notably, our model surpasses these baselines across \textit{goal dist.}, \textit{contact}, \textit{quality}, and \textit{action} scores, signaling a pronounced advancement in generating human motion that aligns accurately with scene and language instructions. Note that the diminished diversity in the \textit{APD} metric mainly stems from the enhanced precision in motion generation within our model, which effectively grounds motions in the 3D scene with desired semantics and interactions, as opposed to the motions with potential physical implausibility, incorrect semantics or inadequate grounding observed in the baselines methods. 

\paragraph{Qualitative Results}

Visualizations in \cref{fig:humanise_qual} reveal that the \ac{cvae} model often fails to accurately ground the target object, indicating limited scene-aware capabilities. Moreover, the one-stage model's lack of scene geometry awareness can result in human-scene collisions and non-contacts.

\paragraph{Affordance Generation Evaluation}

Grounding distance metrics, detailed in \cref{tab:affordance_quan}, illustrate that among the three variants, the MLP lags in grounding accuracy when compared to the Perceiver and Point Transformer models. This discrepancy might arise from the MLP's isolated processing of individual scene points, which limits their information exchange. In contrast, the Perceiver consistently excels, presumably due to its effective integration of point and language features through cross-attention mechanisms.

\begin{table}[ht!]
    \centering
    \small
    \setlength{\tabcolsep}{3pt}
    \caption{\textbf{Quantitative results of affordance map generation.} We report the three distance metrics to evaluate the grounding accuracy.}
    \label{tab:affordance_quan}
        \begin{tabular}{cccc}%
            \toprule
            Arch. of \ac{adm}   & min dist. $\downarrow$ & pelvis dist. $\downarrow$ & all dist. $\downarrow$\\
            \midrule
            G.T.                &  $0.736$           & $0.923$           & $1.039$           \\
            \midrule
            MLP                 &  $0.904^{\pm.003}$ & $1.335^{\pm.008}$ & $1.513^{\pm.008}$ \\
            Point Trans.        &  $0.878^{\pm.008}$ & $1.090^{\pm.008}$ & $1.204^{\pm.009}$ \\
            Perceiver           &  $\mathbf{0.756^{\pm.007}}$ & $\mathbf{1.005^{\pm.005}}$ & $\mathbf{1.086^{\pm.007}}$ \\
            \bottomrule
        \end{tabular}%
\end{table}

\subsection{Results on Novel Evaluation Set}

\begin{table*}[ht!]
    \centering
    \small
    \setlength{\tabcolsep}{3pt}
    \caption{\textbf{Qualitative results on our novel evaluation set.} ``Real'' indicates that we compute these metrics as a reference using the language-motion pairs within the test set of HumanML3D. Of note, our novel evaluation set does not contain ground truth motions.}
    \label{tab:generalize_quan}
    \resizebox{\linewidth}{!}{%
        \begin{tabular}{cccccccccc}%
            \toprule
            Model & \makecell{R-Precision \\ (Top 3)$\uparrow$} & FID$\downarrow$ & \makecell{MultiModal \\ Dist.$\downarrow$} & Diversity$\rightarrow$ & MultiModality$\uparrow$ & contact$\uparrow$ & non-collision$\uparrow$ & \makecell{quality \\ score$\uparrow$} & \makecell{action \\ score$\uparrow$} \\
            \midrule
            Real             & $0.875^{\pm.002}$         & $0.000^{\pm.000}$           & $3.342^{\pm.004}$          & $9.442^{\pm.301}$          & -                          & -                           & -                          & -                      & -                      \\
            \midrule
            one-stage @ Enc & $\mathbf{0.500^{\pm.044}}$ & $11.848^{\pm1.634}$         & $\mathbf{5.954^{\pm.235}}$ & $\mathbf{8.395^{\pm.850}}$ & $4.966^{\pm.321}$          & $46.64^{\pm4.024}$          & $99.88^{\pm.018}$          & $1.94\pm1.15$          & $2.61\pm1.45$          \\
            one-stage @ Dec & $0.403^{\pm.044}$          & $12.268^{\pm.900}$          & $6.611^{\pm.227}$          & $8.049^{\pm.708}$          & $5.031^{\pm.423}$          & $26.75^{\pm4.264}$          & $\mathbf{99.93^{\pm.023}}$ & $1.44\pm0.83$          & $1.96\pm1.27$          \\
            \midrule
            Ours @ Enc      & $0.478^{\pm.069}$          & $\mathbf{7.887^{\pm1.189}}$ & $6.226^{\pm.261}$          & $7.935^{\pm.857}$          & $\mathbf{5.159^{\pm.356}}$ & $71.98^{\pm2.542}$          & $99.83^{\pm.006}$          & $\mathbf{2.06\pm1.23}$ & $\mathbf{2.63\pm1.47}$ \\
            Ours @ Dec      & $0.428^{\pm.023}$          & $12.027^{\pm3.164}$         & $6.412^{\pm.204}$          & $7.603^{\pm.715}$          & $4.966^{\pm.353}$          & $\mathbf{88.63^{\pm2.975}}$ & $99.82^{\pm.015}$          & $1.99\pm1.24$          & $2.49\pm1.40$          \\
            \bottomrule
        \end{tabular}%
    }%
\end{table*}

\begin{figure*}[ht!]
    \centering
    \includegraphics[width=\linewidth]{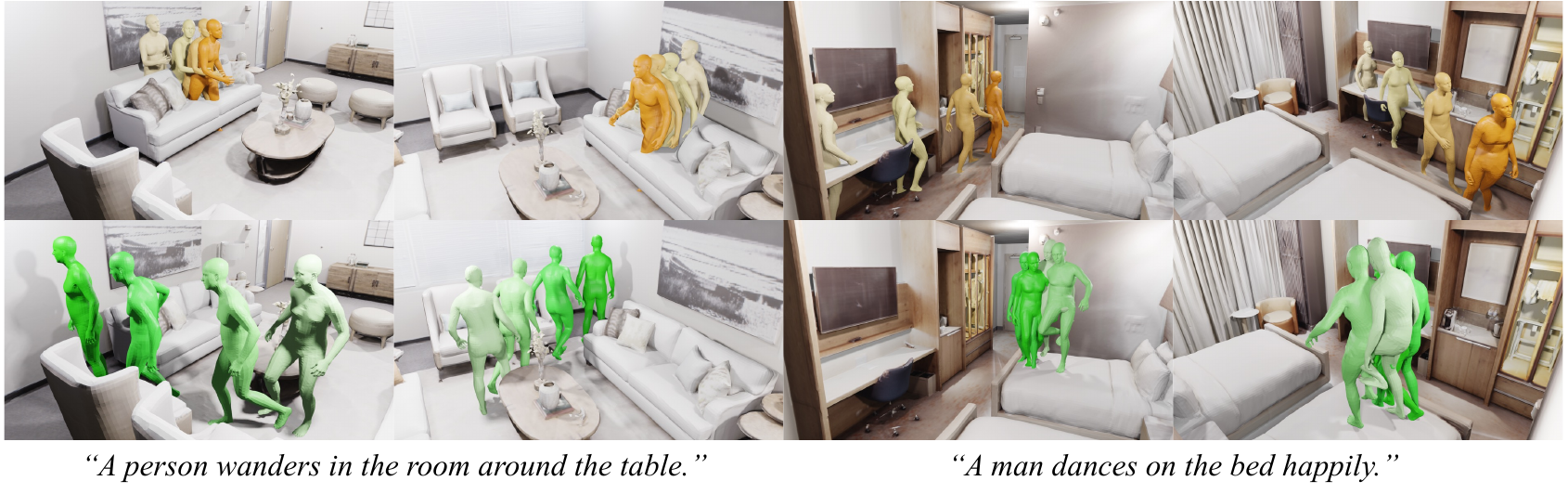}
    \caption{\textbf{Qualitative comparisons on generalization evaluation set.} The first row is generated by the one-stage diffusion model and the second row is generated by our model. Our method can generate natural and accurately grounded human motions in unseen 3D scenes.}
    \label{fig:generalize_qual}
\end{figure*}

\begin{figure}[ht!]
    \centering
    \includegraphics[width=\linewidth]{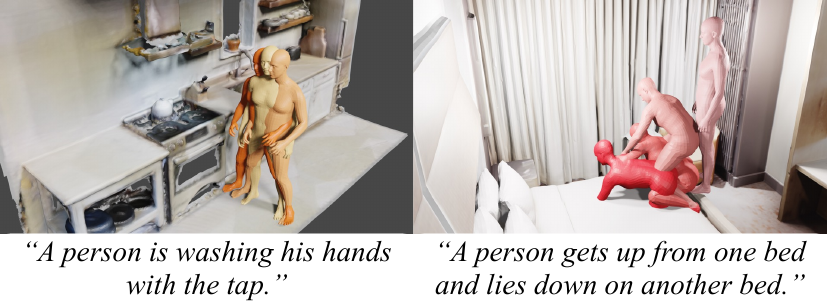}
    \caption{\textbf{Failure cases.} Our model fails while facing entirely unfamiliar \acp{hsi} or too complex descriptions.}
    \label{fig:failure_cases}
\end{figure}

\paragraph{Results}

The performance on the uniquely curated evaluation set, featuring new scenes and language descriptions from Turkers, is summarized in \cref{tab:generalize_quan}. Our approach exhibits considerable enhancements in \textit{FID} while maintaining comparable \textit{R-Precision} and \textit{Multimodal-Dist}, suggesting a robust capability to synthesize plausible human motions aligned with the given language instructions. Notably, our method generates a wider variety of human motions, as evidenced by improved scores in metrics
\textit{MultiModality} when compared to baseline methods. These results underscore our approach's efficacy in producing physically plausible and semantically consistent human motions conditioned on scenes and language instructions, validated through \textit{contact}, \textit{quality}, and \textit{action} scores.
\cref{fig:generalize_qual} presents qualitative results generated with unseen language descriptions and 3D scenes.

\paragraph{Failure Cases}

\cref{fig:failure_cases} depicts typical failure cases encountered by our model. For instance, challenges arise with test scenarios of unseen human-scene interactions, resulting in accurately generated motions in the correct space (\eg, hand washing near a tap) but inaccurate interactions (\eg, failure to align the body appropriately facing the sink). 
The model also fails with language descriptions of high complexity exceeding its current capabilities.

\subsection{Ablation Study}

We further examine the impact of different \ac{adm} architectures on motion generation in the second stage of HUMANISE, utilizing an encoder-based \ac{amdm}. As outlined in \cref{tab:ablation_arch}, both Perceiver and Point Transformer yield superior \textit{goal dist.} outcomes compared to the MLP, echoing findings from \cref{tab:affordance_quan}. Furthermore, these architectures enhance the physical realism, as indicated by improved \textit{contact} scores, with Perceiver models having higher collision rates relative to Point Transformers, echoing observations in \cref{fig:humanise_qual}.

\begin{table}[t!]
    \centering
    \small
    \setlength{\tabcolsep}{3pt}
    \caption{\textbf{Ablation of the architectures of \ac{amdm}.} The Perceiver architecture slightly outperforms the Point Transformer in the metrics of \textit{goal dist.} and \textit{contact} score.}
    \label{tab:ablation_arch}
        \begin{tabular}{cccc}%
            \midrule
            Arch. of \ac{adm} & goal dist.$\downarrow$ & contact$\uparrow$ & non-collision$\uparrow$ \\
            \midrule
            G.T.         & $0.017$                      & $90.79$                       & $99.84$                    \\
            \midrule
            MLP          & $0.394^{\pm.010}$            & $73.96^{\pm.434}$             & $\mathbf{99.84^{\pm.005}}$ \\
            Point Trans. & $0.164^{\pm.010}$            & $94.39^{\pm.408}$             & $99.82^{\pm.008}$          \\
            Perceiver    & $\mathbf{0.156^{\pm.006}}$   & $\mathbf{95.86^{\pm.323}}$    & $99.69^{\pm.007}$          \\
            \bottomrule
        \end{tabular}%
\end{table}

\section{Conclusion}

We introduced a novel two-stage model that leverages scene affordance as an intermediate representation to bridge the 3D scene grounding and subsequent conditional motion generation. The quantitative and qualitative results demonstrate promising improvements in HumanML3D and HUMANISE. The model's adaptability was further validated on a uniquely curated evaluation set featuring unseen scenes and language prompts, showcasing its robustness in novel scenarios.

\paragraph{Limitations}
(i) The reliance on diffusion models contributes to slower inference times, marking a significant drawback for future work. (ii) Although employing affordance maps mitigates the challenges posed by the scarcity of paired data for training in 3D environments, data limitation remains a critical hurdle. Future initiatives should focus on devising strategies to overcome this persistent challenge.

\paragraph{Acknowledgments}

The authors would like to thank NVIDIA for their generous support of GPUs and hardware. This work is supported in part by the National Science and Technology Major Project (2022ZD0114900), the National Natural Science Foundation of China (NSFC) (62172043), and the Beijing Nova Program.

{\small
\bibliographystyle{ieeenat_fullname}
\bibliography{reference_header,reference}
}

\clearpage
\appendix
\renewcommand\thefigure{A\arabic{figure}}
\setcounter{figure}{0}
\renewcommand\thetable{A\arabic{table}}
\setcounter{table}{0}
\renewcommand\theequation{A\arabic{equation}}
\setcounter{equation}{0}
\pagenumbering{arabic}%
\renewcommand*{\thepage}{A\arabic{page}}
\setcounter{footnote}{0}
\setlength\floatsep{1\baselineskip plus 3pt minus 2pt}
\setlength\textfloatsep{1\baselineskip plus 3pt minus 2pt}
\setlength\dbltextfloatsep{1\baselineskip plus 3pt minus 2 pt}
\setlength\intextsep{1\baselineskip plus 3pt minus 2 pt}

\section{SMPL-X Joints to Meshes} \label{app:sec:joints2mesh}

We represent the per-frame human pose using SMPL-X \citep{pavlakos2019expressive} body joints, denoted as $\bx_i \in \mathbb{R}^{J \times 3}$. Among the available 55 SMPL-X joints, we utilize $J=22$ joints, excluding those related to the hands, jaw, and eyes. To facilitate visualization and compute physical metrics, we transform the SMPL-X joint positions of motion sequences into meshes through a two-stage optimization process.
In the first stage, we employ a pre-trained transformer-based neural network to map the joint sequence to the corresponding SMPL-X parameter sequence, providing an initialization for the subsequent stage.
Then, in the second stage, we optimize the SMPL-X parameters using an MSE loss function, aiming to minimize the discrepancy between joints derived from the optimized SMPL-X parameter and the generated joints.

\section{Model Architectures} \label{app:sec:architecture}

\subsection{Details of \ac{adm} and \ac{amdm}}

\paragraph{\ac{adm}} We first employ a pre-trained Point Transformer \citep{zhao2021point}, which is pre-trained on semantic segmentation task, to extract per-point features from the input 3D scene. We then concatenate the noisy affordance map, per-point feature, and point coordinates; forward the concatenation into the backbone of \ac{adm}.
We use $J=6$ to compute the affordance map and extract per-point features with a dimensionality of 32, containing joints of the pelvis, left/right hand, left/right foot, and neck.

\begin{figure*}[t!]
    \centering
    \includegraphics[width=\linewidth]{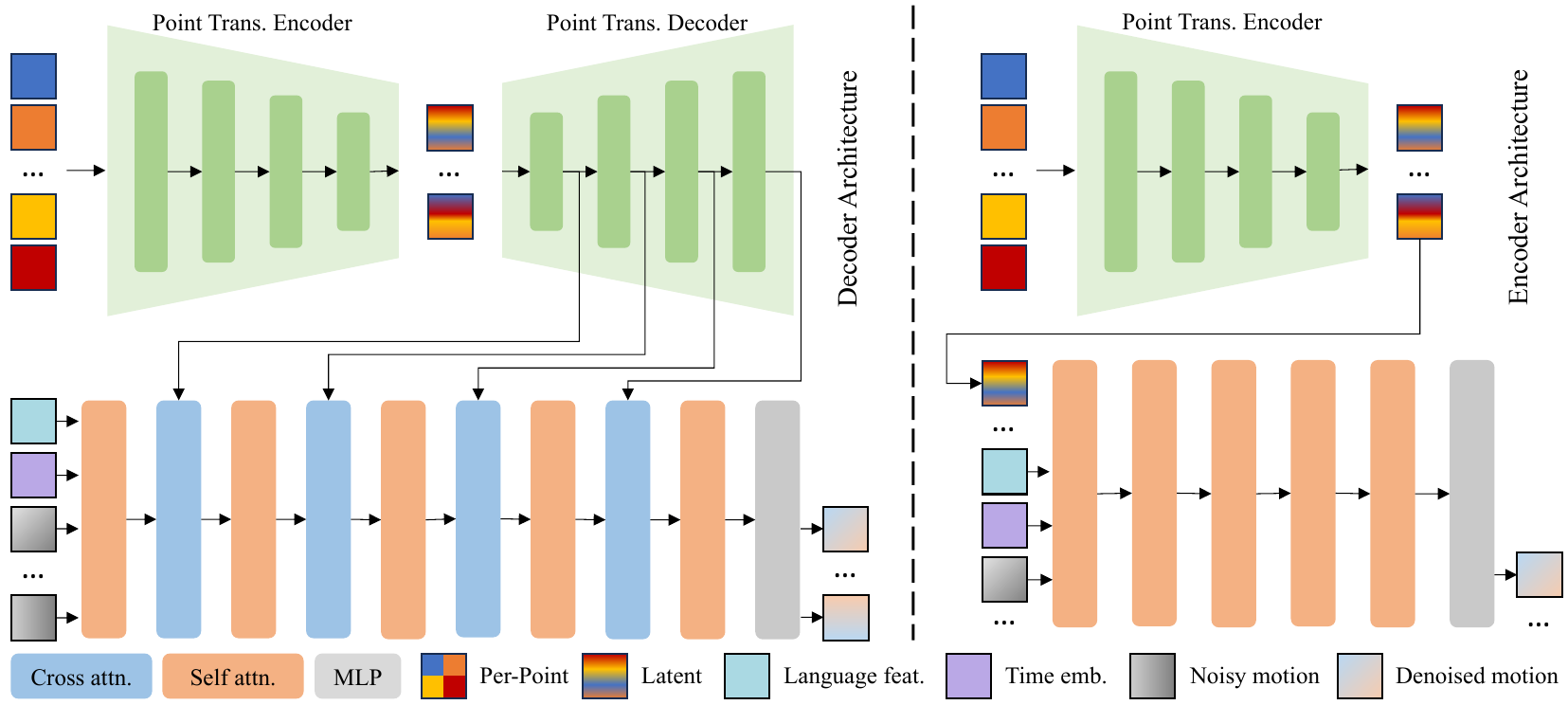}
    \caption{\textbf{Illustration of the decoder and encoder variants' architectures.} The left part depicts the architecture of the decoder variant, which stacks self-attention and cross-attention layers alternately to fuse multi-modal conditions effectively. The right part showcases the design of the encoder variant, employing self-attention layers to fuse the language features, affordance features, and noisy motion sequences.}
    \label{fig:supp:dec_enc}
\end{figure*}

\paragraph{\ac{amdm}} For \ac{amdm}, we have implemented both a decoder and an encoder variant for comparison. The main difference between the two architectures is the approach used for fusing affordance features.
As shown in \cref{fig:supp:dec_enc}, the decoder variant utilizes cross-attention to attend with affordance features, while the encoder variant directly employs self-attention for the fusion process.
We utilize the Pytorch implementation of \texttt{TransformerEncoderLayer} and \texttt{TransformerDecoderLayer}, configuring the latent space dimension to be 512.

\subsection{Architecture of \ac{adm} Variants}

\begin{figure*}[t!]
    \centering
    \begin{subfigure}{0.4\linewidth}
        \includegraphics[width=\linewidth]{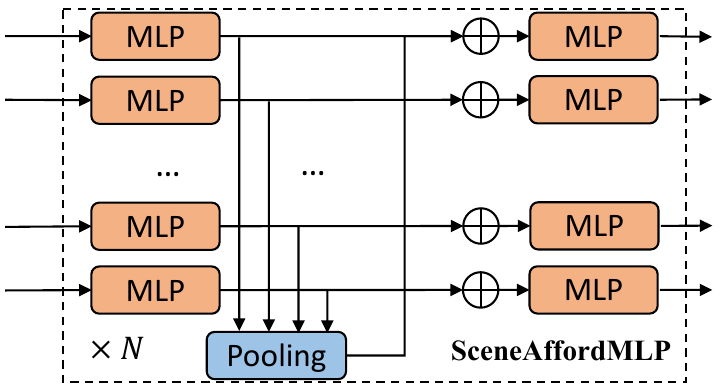}
        \caption{MLP variant}
        \label{fig:supp:mlp}
    \end{subfigure}%
    \begin{subfigure}{0.6\linewidth}
        \includegraphics[width=\linewidth]{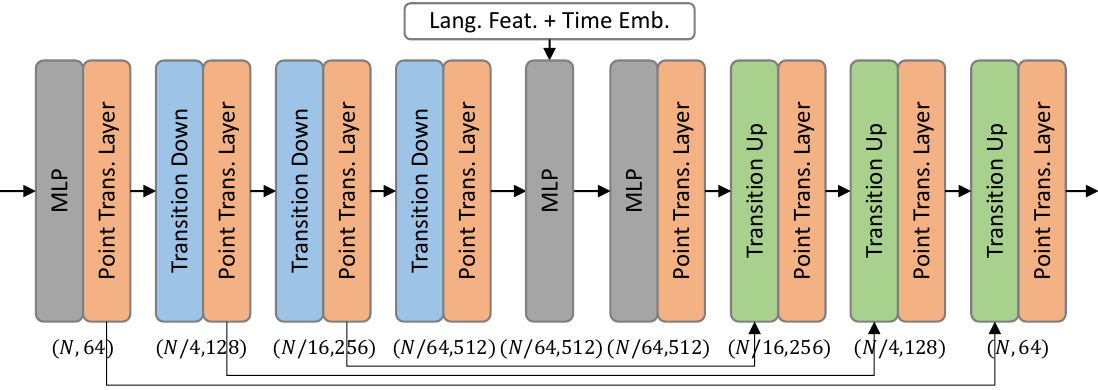}
        \caption{Point Transformer variant}
        \label{fig:supp:pointtrans}
    \end{subfigure}%
    \caption{\textbf{Illustration of MLP and Point Transformer variants of \ac{adm}.}}
    \label{fig:supp:adm_variants}
\end{figure*}

\paragraph{MLP}
The MLP variant consists of two ``SceneAffordMLP'' blocks, as depicted in \cref{fig:supp:mlp}.
In each ``SceneAffordMLP'' block, we first process the input via a shared MLP layer. Subsequently, a max-pooling layer aggregates information from all points to extract the global feature. The per-point features are then concatenated with the global feature and fed into another shared MLP layer. This model directly operates on the concatenation of per-point features, language features, timestep embedding, and noisy affordance.

\paragraph{Point Transformer} 
The Point Transformer variant employs a U-Net architecture that comprises an encoder and a decoder for extracting point-wise features, as shown in \cref{fig:supp:pointtrans}.
The feature encoder comprises four stages designed to gradually downsample the point cloud with transition-down blocks and aggregate point features with point transformer layers. The downsampling rates at each stage are specified as $\left[1, 4, 4, 4\right]$, resulting in point cloud cardinalities of $\left[N, N/4, N/16, N/64\right]$ after each stage.
The decoder, incorporating transition-up blocks and point transformer layers, maps the encoded features onto a higher-resolution point set. This is achieved by concatenating the trilinear-interpolated features from the previous decoder stage with features from the corresponding encoder stage through a skip connection. To enable language-conditioned modeling, we enhance this connection by incorporating a linear layer that fuses language and point features. Ultimately, the per-point feature vectors are forwarded into a linear layer.

\section{Implementation Details} \label{app:sec:implementation}

\subsection{Baseline Models for Motion Generation}

\paragraph{cVAE} We adopt the model architecture and hyperparameters as proposed by \citet{wang2022humanise} without any alterations. We avoid utilizing the suggested auxiliary loss functions to ensure a fair comparison.

\paragraph{One-stage Baselines} Our one-stage baselines directly adopt the architecture illustrated in \cref{fig:supp:dec_enc}. Unlike \ac{amdm}, the models take input as the scene point coordinates and semantic features instead of the affordance map. We extract the per-point semantic features using a Point Transformer, which is pre-trained on a semantic segmentation task.

\subsection{Training Details}

In all experiments, we fix the diffusion step of \ac{adm} as 500.
In the experiments on the HumanML3D dataset, we set the diffusion step of \ac{amdm} as 1000, following the MDM \citep{tevet2023human}.
Given the presence of only a floor within the scene in HumanML3D, we opt not to utilize the pre-trained Point Transformer to extract semantic features in \ac{adm}.
\textbf{To mitigate overfitting of the generated results to ground truth scene affordances, we randomly replace half of the ground truth affordances (\ie, $50\%$ proportion) with predicted ones.} We conduct an ablative experiment on the proportion and report the results in \cref{app:tab:ablation_mix_train}.
For the HUMANISE benchmark, we set the diffusion step of \ac{amdm} as 500, and we directly use the ground truth affordance map during the training in the second stage. We augment each scene point cloud by applying random rotation around the z-axis.
For the experiments on our novel evaluation set, we set the diffusion step of \ac{adm} and \ac{amdm} as 500 and 1000, respectively. We directly utilize the ground truth affordance map in the second-stage training. \cref{app:tab:ablation_mix_train_novel_set} shows the ablative results of replacing the ground truth affordance with predicted ones using different proportions.

\subsection{Data Preprocessing}

We curate a dataset that connects language, 3D scene, and motion by incorporating data from several sources, \ie, HUMANISE, HumanML3D, and PROX. For HUMANISE and PROX containing scene context, we crop the scene into $4 \times 4 \text{m}^2$ chunks according to the motion range. For HumanML3D, we re-process the data without normalizing the orientation of the first frame pose and randomly add the floor and $3 \sim 4$ furniture scans around the motion sequence. We use scans from ScanObjectNN \citep{uy2019revisiting}, including the categories of ``table'', ``chair'', ``bed'', ``desk'' ``sofa'' ``shelf'', ``door'', and ``toilet''. For each scene, we downsample the point cloud into 8192 points.

\subsection{User Study}

We performed human perceptual studies to assess the generated results on both HUMANISE and our novel evaluation set. We randomly generated 20 samples for each model and presented them in a disarranged manner. We asked 12 workers to score each sample on a scale from $1$ to $5$ for the \textit{quality} and \textit{action} score. A higher score indicates a better result. The \textit{quality} score reflects the overall quality of the generation, while the \textit{action} score evaluates the consistency of the generated motions with the provided descriptions.

\section{More Qualitative Results} \label{app:sec:more_qual_results}

\begin{figure*}[t!]
    \centering
    \includegraphics[width=\linewidth]{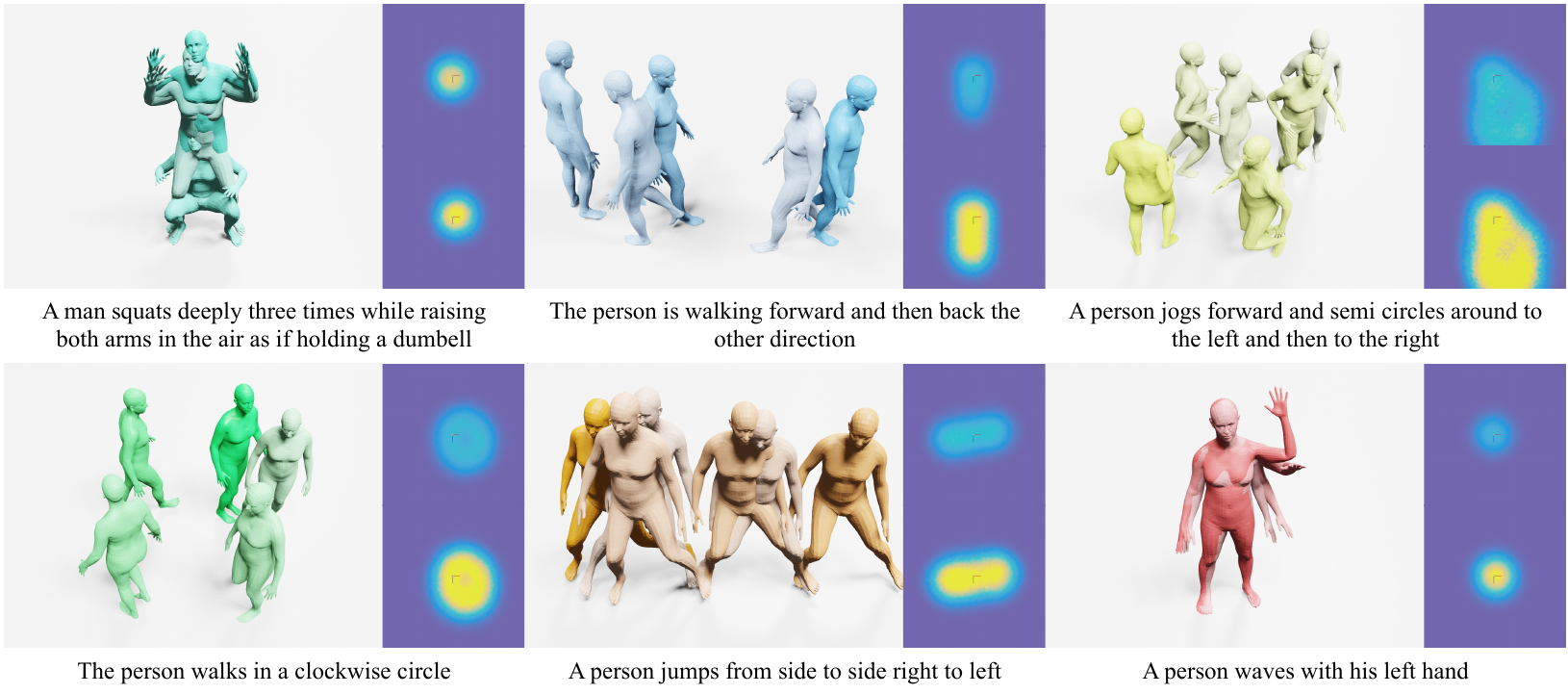}
    \caption{\textbf{Qualitative results on HumanML3D.} In each case, the left figure illustrates the generated motion and the generated affordance maps are depicted in the two figures on the right. The top and bottom figures correspond to the pelvis and left foot joints, respectively.}
    \label{fig:supp:results_h3d}
\end{figure*}

We present the qualitative results on the HumanML3D dataset in \cref{fig:supp:results_h3d}. Please visit our \href{https://afford-motion.github.io}{project page}, where you can find rendered videos showcasing more qualitative results.

\begin{table*}[t!]
    \centering
    \small
    \setlength{\tabcolsep}{3pt}
    \caption{\textbf{Ablation of the proportion about replacing ground truth affordance with predicted ones on HumanML3D.} $0\%$ indicates we train \ac{amdm} using ground truth affordance, $100\%$ indicates we use the predicted affordance, and $50\%$ indicates we randomly replace half of the ground truth affordance with predicted ones. We use the Perceiver in the first stage and the encoder-based variant in the second stage.}
    \label{app:tab:ablation_mix_train}
        \begin{tabular}{cccccccc}%
            \toprule
            \multirow{2}{*}{Proportion} & \multicolumn{3}{c}{R-Precision $\uparrow$} & \multirow{2}{*}{FID $\downarrow$} & \multirow{2}{*}{MultiModal Dist. $\downarrow$} & \multirow{2}{*}{Diversity $\rightarrow$} & \multirow{2}{*}{MultiModality $\uparrow$}\\
            \cline{2-4} & Top 1 &  Top 2 & Top 3 & \\
            \midrule
            Real    & $0.511^{\pm.003}$ & $0.703^{\pm.003}$ & $0.797^{\pm.002}$ & $0.002^{\pm.000}$ & $2.974^{\pm.008}$ & $9.503^{\pm.065}$ & - \\
            \midrule
            $0\%$   & $0.291^{\pm.014}$ & $0.434^{\pm.013}$ & $0.528^{\pm.016}$ & $2.482^{\pm.477}$ & $4.784^{\pm.103}$ & $8.986^{\pm.103}$ & $4.123^{\pm.046}$ \\
            $50\%$  & $0.432^{\pm.007}$ & $0.629^{\pm.007}$ & $0.733^{\pm.006}$ & $0.352^{\pm.109}$ & $3.430^{\pm.061}$ & $9.825^{\pm.159}$ & $2.835^{\pm.075}$ \\
            $100\%$ & $0.415^{\pm.010}$ & $0.599^{\pm.013}$ & $0.703^{\pm.010}$ & $0.537^{\pm.218}$ & $3.574^{\pm.078}$ & $9.730^{\pm.093}$ & $3.241^{\pm.042}$ \\
            \bottomrule
        \end{tabular}%
\end{table*}

\begin{table*}[t!]
    \centering
    \small
    \setlength{\tabcolsep}{3pt}
    \caption{\textbf{Ablation of the proportion about replacing ground truth affordance with predicted ones on our novel evaluation set.} The proportion ranges from $0.0$ to $0.5$. We use the Perceiver in the first stage and the encoder-based variant in the second stage.}
    \label{app:tab:ablation_mix_train_novel_set}
    \resizebox{\linewidth}{!}{%
        \begin{tabular}{cccccccccc}%
            \toprule
            \multirow{2}{*}{Proportion} & \multicolumn{3}{c}{R-Precision $\uparrow$} & \multirow{2}{*}{FID $\downarrow$} & \multirow{2}{*}{MultiModal Dist. $\downarrow$} & \multirow{2}{*}{Diversity $\rightarrow$} & \multirow{2}{*}{MultiModality $\uparrow$} & \multirow{2}{*}{contact$\uparrow$} & \multirow{2}{*}{non-collision$\uparrow$} \\
            \cline{2-4} & Top 1 &  Top 2 & Top 3 & \\
            \midrule
            Real   & $0.588^{\pm.006}$ & $0.784^{\pm.003}$ & $0.875^{\pm.002}$ & $0.000^{\pm.000}$ & $3.342^{\pm.004}$ & $9.442^{\pm.301}$ & -       & -       & -\\
            \midrule
            $0\%$  & $0.205^{\pm.054}$ & $0.343^{\pm.056}$ & $0.478^{\pm.069}$ & $7.887^{\pm1.189}$& $6.226^{\pm.261}$ & $7.935^{\pm.857}$ & $5.159^{\pm.356}$ & $71.98^{\pm2.542}$& $99.83^{\pm.006}$ \\
            $30\%$ & $0.238^{\pm.017}$ & $0.358^{\pm.023}$ & $0.488^{\pm.026}$ & $11.457^{\pm1.219}$&$5.896^{\pm.109}$ & $8.012^{\pm.378}$ & $4.786^{\pm.249}$ & $34.03^{\pm.661}$ & $99.89^{\pm.024}$ \\
            $50\%$ & $0.253^{\pm.037}$ & $0.415^{\pm.045}$ & $0.500^{\pm.042}$ & $13.354^{\pm1.281}$&$5.747^{\pm.203}$ & $7.976^{\pm.487}$ & $4.649^{\pm.337}$ & $30.54^{\pm2.296}$& $99.92^{\pm.016}$ \\
            \bottomrule
        \end{tabular}%
    }%
\end{table*}

\section{Novel Evaluation Set} \label{app:sec:novel_eval}

\begin{figure*}[t!]
    \centering
    \includegraphics[width=\linewidth]{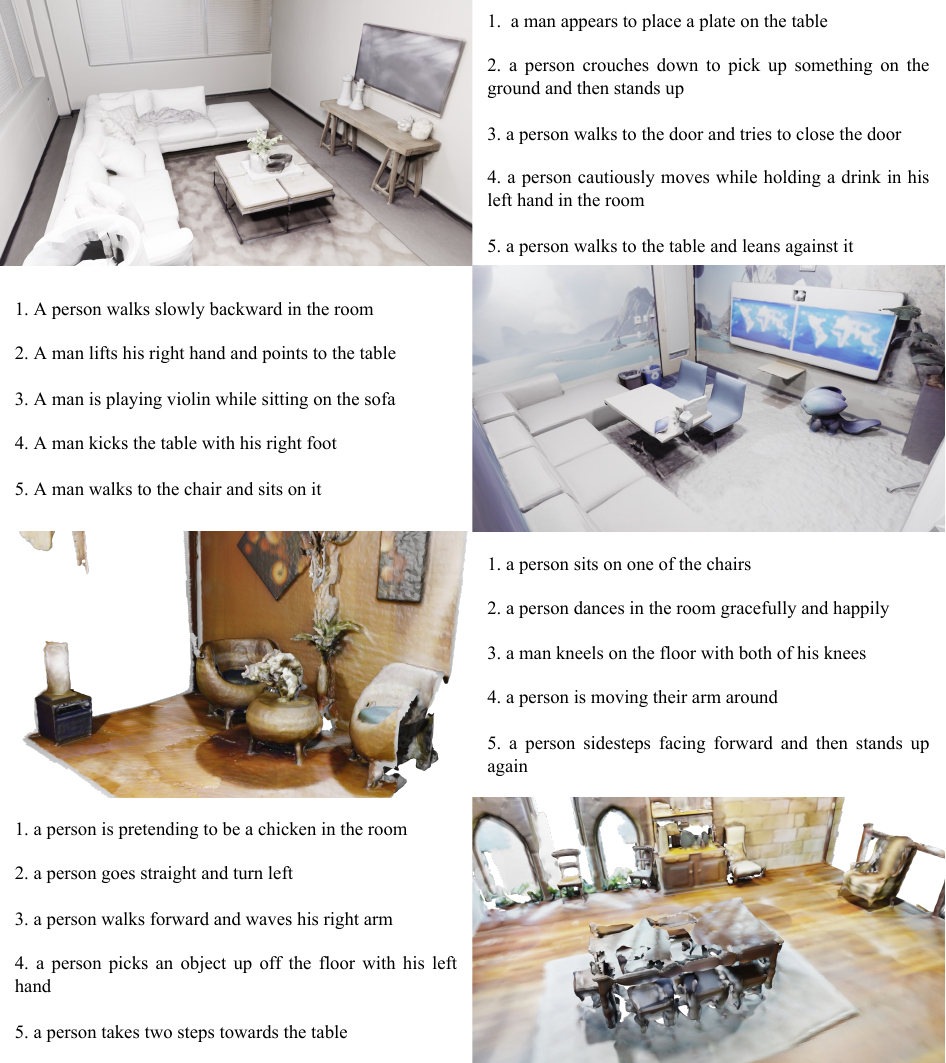}
    \caption{\textbf{Example cases in novel evaluation set.}}
    \label{fig:supp:novel_eval}
\end{figure*}

We establish a novel evaluation set where both the scene and language descriptions are unseen during the training. We visualize some evaluation cases in \cref{fig:supp:novel_eval}.

\paragraph{Evaluation Details} Employing joint position representation, we re-split the HumanML3D dataset and follow \citet{guo2022generating} to utilize the training set to train feature extractors, which supports the computation of metrics like \textit{R-Percision}, \textit{FID}, \textit{MultiModal Dist.}, \textit{Diversity}, and \textit{MultiModality}. For \textit{R-Precision} computation, we form a description pool with one ground truth and 15 randomly selected mismatched descriptions. To evaluate the model's performance on our established novel evaluation set, we use the HumanML3D test set to compute these metrics as a reference (\ie, ``Real''). It's worth noting that our evaluation set lacks ground truth motion and differs from the HumanML3D dataset regarding description distribution, \ie, utterances in HumanML3D include fewer scene-related descriptions.

\end{document}